\documentclass[10pt,twocolumn,letterpaper]{article}
\usepackage[pagenumbers]{cvpr}
\usepackage{graphicx}
\usepackage{stfloats}
\usepackage{xcolor}
\usepackage{booktabs}
\usepackage{colortbl}
\usepackage{amsmath}
\usepackage{amssymb}
\usepackage{amsfonts}
\usepackage{mathtools}
\usepackage{bm}
\usepackage{pifont}
\usepackage{nicefrac}
\usepackage{microtype}
\usepackage{xspace}
\usepackage{multirow}
\usepackage{arydshln}
\usepackage{algorithm}
\usepackage{algpseudocode}
\definecolor{cvprblue}{rgb}{0.21,0.49,0.74}
\definecolor{tcrrowblue}{RGB}{235,244,252}

\newcommand{\tcrpct}[1]{\textcolor{cvprblue}{(#1\%)}}
\newcommand{\metricheading}[3][7.5]{{\fontsize{#1}{9}\selectfont\bfseries\shortstack{#2\\#3}}}

\usepackage[breaklinks,colorlinks,allcolors=cvprblue]{hyperref}
\newcommand{\projectlink}{\href{https://lyongo.github.io/PhyS/}{Project Page}}

\def\onedot{\xspace}

\def\name{PhyS\xspace}

\title{Distilling Physical Priors into Streaming World Models}
\author{
Liangliang Zhao\textsuperscript{\rm 1,2},
Junying Wang\textsuperscript{\rm 2},
Danni Yang\textsuperscript{\rm 2},
Yifan Chang\textsuperscript{\rm 2},\\
Bin Fu\textsuperscript{\rm 2},
Yu Qiao\textsuperscript{\rm 2},
Bowen Zhou\textsuperscript{\rm 2},
Yihao Liu\textsuperscript{\rm 2}\thanks{Corresponding author.}\\
\textsuperscript{\rm 1}Fudan University, Shanghai, China\\
\textsuperscript{\rm 2}Shanghai AI Laboratory, Shanghai, China\\
{\tt\small\{zhaoliangliang, liuyihao\}@pjlab.org.cn}
}

\begin{document}
\maketitle
\begin{abstract}
Streaming world models predict future visual states through online rollouts while maintaining physically coherent dynamics over long horizons. However, their rollouts often violate basic physical constraints. A common approach builds such models by distilling pretrained bidirectional DiTs into few-step causal generators. However, this paradigm suffers from two fundamental limitations: generic bidirectional teachers acquire limited physical priors from visually oriented pretraining, and the limited priors suffer further loss during bidirectional-to-causal distillation. We present \name, a three-stage framework for distilling physical priors into streaming world models. To acquire physical priors from real-world interactions, we construct \name-120K, a dataset of 120K real-world physical-interaction videos spanning rigid-body dynamics, soft-body deformation, fluid phenomena, and phase transitions. Each video is annotated with structured descriptions of object properties and causal state transitions. Physics-aware supervised fine-tuning injects the physical priors into a bidirectional 14B DiT teacher, which we then distill into a lightweight 1.3B causal DiT for few-step autoregressive streaming generation. Finally, we use online reinforcement learning to incentivize the distilled model to generate physically plausible rollouts and further propose Temporal Credit Routing (TCR) to address temporal credit assignment. TCR evaluates physical consistency over overlapping temporal windows and routes the resulting group-relative advantages to temporally aligned denoising actions. On PhysicsIQ, fine-tuning improves Wan2.1-14B by 18.2\%, and the full \name pipeline outperforms the original Self Forcing, Rolling Forcing, and Causal Forcing baselines by 23.7\%, 14.8\%, and 31.4\%, respectively. The results further show improvements on the physics-aware video generation benchmarks of VideoPhy, VideoPhy2, and PhyGenBench. Visit our \projectlink{}.
\end{abstract}

\begin{figure}[!t]
  \centering
  \includegraphics[width=\linewidth]{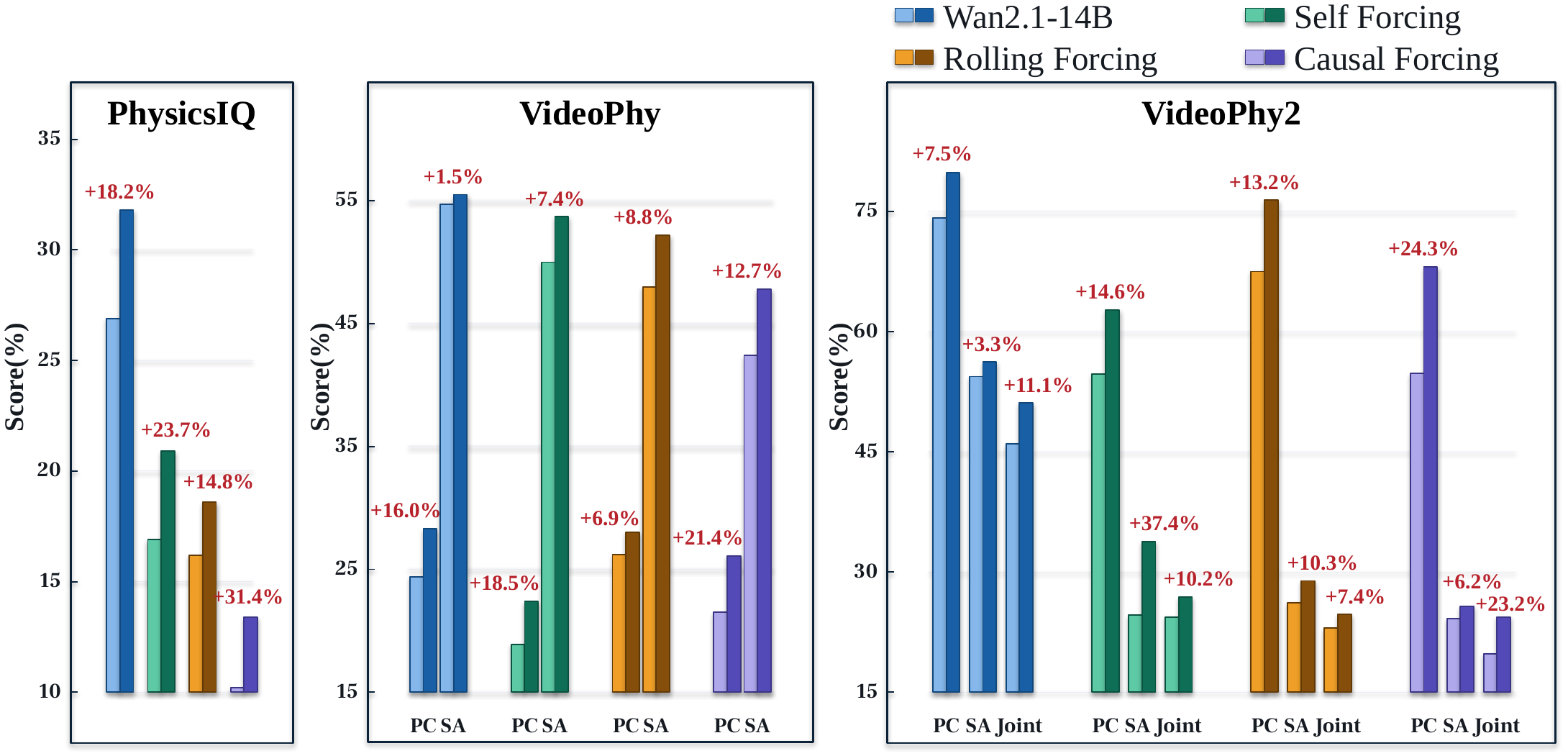}
  \caption{Benchmark gains. Light bars show baselines. Dark bars show the SFT teacher or Forcing models with full \name.}
  \label{fig:benchmark_improvements}
\end{figure}

\begin{figure*}[!t]
  \centering
  \includegraphics[width=0.85\linewidth]{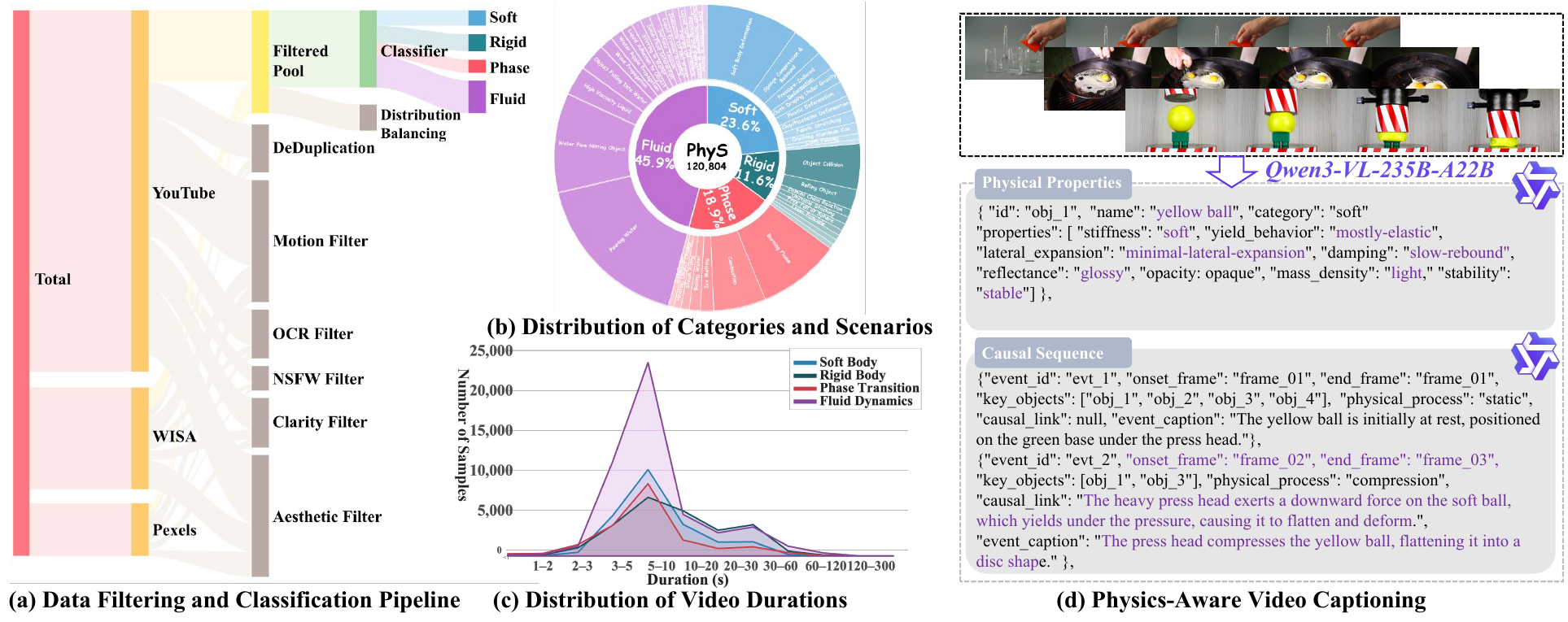}
  \caption{
  Overview of the \name-120K construction pipeline and dataset.
  (a) Raw clips from YouTube, Pexels, and WISA are filtered,
  classified, and subsequently rebalanced to limit the largest
  category share.
  (b) Distribution of physical-process categories and scenarios.
  (c) Most clips fall between 3--10\,s, with the largest bin at
  5--10\,s.
  (d) Structured physical properties and causal event sequences
  are generated by Qwen3-VL and refined through random human inspection.
  }
  \label{fig:dataset_overview}
\end{figure*}

\section{Introduction}
\label{sec:intro}

In recent years, video generation models~\citep{videoworldsimulators2024,polyak2024movie,ma2025step,wan2025wan} have made significant progress and can now synthesize videos with high visual fidelity and temporal coherence. Large-scale video pre-training enables these models to learn rich spatiotemporal representations, motion patterns, and scene dynamics, laying a crucial foundation for building video world models. Building on these learned priors, streaming world models can predict future visual states from initial observations, thereby representing world dynamics in video form.
Given observed context frames and an optional text condition, they generate future frames block by block, conditioning each block on the previously generated history. This causal formulation enables continuous streaming rollout without regenerating the complete video sequence.

Existing methods commonly adopt a Forcing-based training paradigm that distills a pretrained bidirectional DiT into a few-step causal generator~\citep{huang2025selfforcing,liu2025rolling,zhu2026causal}. This paradigm enables streaming world models to perform long-horizon rollouts, yet their predictions frequently violate fundamental physical constraints, causal relationships, and commonsense world knowledge. This problem arises from two fundamental limitations of the paradigm. First, generic bidirectional teachers acquire limited physical priors from visually oriented pretraining. Second, these limited priors suffer further loss during bidirectional-to-causal distillation, making it difficult for the causal student to preserve the long-range dependencies essential to physical causality. The resulting errors accumulate throughout autoregressive rollouts, exacerbating physical and causal deviations over long horizons.

To address these limitations, we propose \textbf{\name}, a training framework for physical streaming world models. We first design a data pipeline and use it to collect and curate \name-120K, a dataset of 120{,}804 real-world physical-interaction videos covering representative physical processes such as rigid-body dynamics, soft-body deformation, fluid dynamics, and phase transitions. We then employ a vision-language model~\citep{bai2025qwen3} to annotate each clip with a structured physical description, providing fine-grained supervisory signals for learning how physical dynamics evolve in real-world scenarios. Using this dataset, we perform physics-aware supervised fine-tuning with a bidirectional 14B DiT video generator as the backbone, yielding a teacher model that captures prior knowledge of real-world physical dynamics. We then distill this 14B teacher into a lightweight 1.3B causal DiT that supports few-step inference, enabling streaming video generation conditioned on contextual frames. Finally, a reward model scores generated videos for physical consistency and guides online optimization of the causal DiT.

The final stage aligns the causal generator through Temporal Credit Routing (TCR). Instead of broadcasting a single video-level advantage, TCR routes rewards across overlapping temporal windows to aligned denoising actions. This overlap-routed signal localizes physical feedback. As shown in Fig.~\ref{fig:benchmark_improvements}, physics-aware SFT improves the Wan2.1-14B teacher's PhysicsIQ score by 18.2\%. The full \name pipeline yields relative PhysicsIQ gains of 23.7\%, 14.8\%, and 31.4\% over the original Self Forcing, Rolling Forcing, and Causal Forcing baselines, respectively, with gains across the reported VideoPhy and VideoPhy2 metrics.

Our contributions are summarized as follows:
\begin{itemize}
\item \textbf{Dataset:} We construct \textbf{\name-120K}, a dataset of 120K real-world physical-interaction videos with structured annotations of object properties and causal state transitions.
\item \textbf{Framework:} We introduce \textbf{\name}, a three-stage framework that learns physical priors with a bidirectional DiT teacher and transfers them to a lightweight, few-step causal generator for streaming rollout.
\item \textbf{Optimization Strategy:} We propose \textbf{Temporal Credit Routing (TCR)}, which aggregates window-level physical rewards into overlap-routed advantages for causal denoising actions and improves multiple causal video generators across four physics-aware benchmarks.
\end{itemize}
\section{Related Work}
\label{sec:related}
\paragraph{\textbf{Autoregressive Video Generation.}}
Modern video diffusion models denoise complete sequences with bidirectional temporal context, whereas streaming world models require causal, low-latency generation; autoregressive (AR) approaches meet this requirement by conditioning each frame or chunk on previously generated content. Early AR methods represent videos as discrete spatiotemporal tokens~\citep{yan2021videogpt,kondratyuk2023videopoet,bruce2024genie}, while recent causal diffusion and flow-matching models operate in latent space using few-step denoising or blockwise generation~\citep{chen2024diffusionforcing,yin2025causvid,huang2025selfforcing,teng2025magi,gu2025long,jin2024pyramidal}. They are trained with causal-teacher or diffusion-forcing objectives~\citep{hu2024acdit,gao2024ca2,song2025history}, or distilled from bidirectional teachers through distribution matching~\citep{yin2024dmd,yin2025causvid,huang2025selfforcing}; rolling-window and history-aware training further improve rollout stability~\citep{ruhe2024rolling,chen2025skyreels,liu2025rolling}.
Beyond rollout stabilization, AR-CoPO~\citep{he2026arcopo} scores each complete branch to update only the forked chunk, while TCR routes overlapping window feedback across blocks by frame overlap without branching. 

\paragraph{\textbf{Physics-aware Video Generation.}}
Physics-aware video generation incorporates physical knowledge through simulation-based, language-guided, and learned-prior approaches. Explicit-model approaches impose dynamics using physical solvers, reconstructed states, or programmatic animation~\citep{stomakhin2013material,jiang2016material,liu2024physgen,li2023pac,chen2025vid2sim,lv2024gpt4motion,li2025wonderplay,xie2025physanimator,liu2026realwonder}, but depend on accurate reconstruction, hand-specified dynamics, or costly simulation. Language-guided methods use LLMs or VLMs to produce physical descriptions or causal rationales~\citep{xue2025phyt2v,zhang2025think,hao2025enhancing,wang2026chain}, but their global signals remain too coarse for frame-level state changes. Learned-prior methods internalize physical knowledge through fine-tuning, representation distillation, local conditioning, expert modules, or feedback-based optimization~\citep{wang2025wisa,videorepa2025,gillman2026goal,lu2026phys4d,pathak2026physvid,wang2026prophy,ji2025physmaster,lin2025reasoning,zhang2026physrvg}.

\section{\name-120K Dataset}
\label{sec:dataset}
We design a process-centric data construction pipeline that automatically curates, organizes, and annotates video clips rich in visible physical processes from large-scale real-world videos. Using this pipeline, we construct \name-120K, a large-scale real-world physical video dataset comprising 120{,}804 video clips. Prior physics-related video datasets and benchmarks mainly rely on synthetic environments, generated prompts, controlled motion patterns, or real-world settings with limited process diversity and scale~\citep{Yi2020,Bear2021,Tung2023,Bansal2025,wang2025wisa,li2025pisa,zhou2026physinone}. \name-120K improves over these resources in real-scene grounding, process diversity, and scale, providing observation-grounded dynamics supervision for physically consistent video world models.

\begin{figure}[t]
  \centering
  \includegraphics[width=\linewidth]{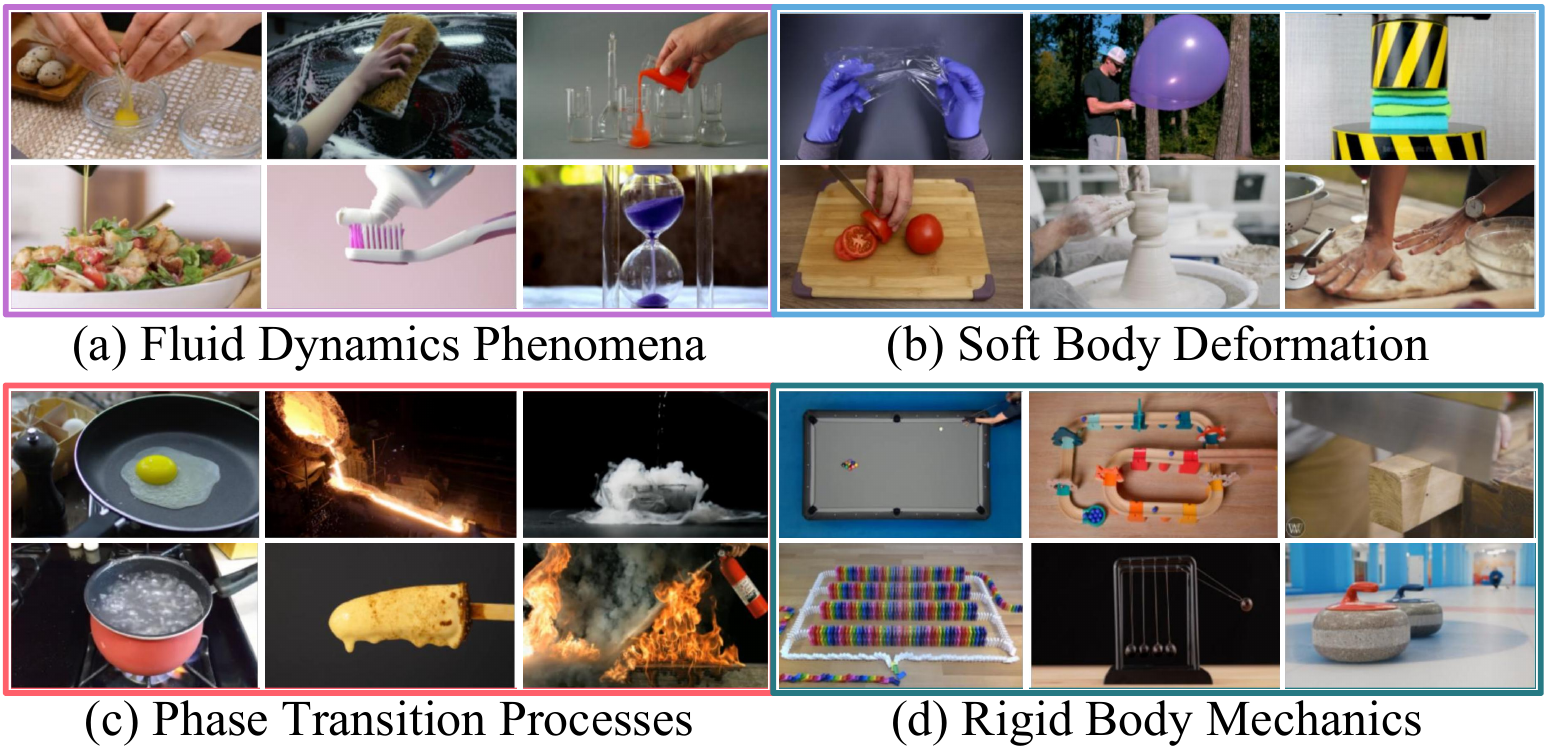}
  \caption{Representative scenarios from \name-120K.}
  \label{fig:dataset_gallery}
\end{figure}

\paragraph{A process-centric taxonomy.}
Current video generation models can synthesize visually realistic frames, yet they often produce physically implausible results in scenes involving physical processes, in large part because they are not fine-tuned on datasets that specifically capture real-world physical dynamics. Therefore, \name-120K organizes clips according to observable state evolution and interaction patterns into four physical-process categories (Fig.~\ref{fig:dataset_gallery}): fluid dynamics (pouring, splashing, diffusion), soft-body deformation (compression, stretching, tearing), phase transitions (melting, boiling, combustion), and rigid-body mechanics (collision, rolling, balancing); the partition is grounded in the physical process itself rather than object identity or isolated physical laws. Figure~\ref{fig:dataset_overview}(b) shows the four categories, their clip shares (45.9\%, 23.6\%, 18.9\%, and 11.6\%, respectively), and the scenarios within each.

\paragraph{Data curation and annotation pipeline.}
Fig.~\ref{fig:dataset_overview}(a) summarizes the automated construction pipeline. We collect candidate videos from public video platforms and open-source datasets, including YouTube, Pexels, and WISA~\citep{wang2025wisa}, using process-oriented queries, and segment long videos into single-shot clips with TransNetV2~\citep{soucek2024transnet}. We then apply near-duplicate removal and multi-stage filtering for motion quality, camera artifacts, visible text, unsafe content, clarity, and aesthetics, using tools such as RAFT~\citep{teed2020raft}, OpenNSFW2~\citep{beikmohammadi2021opennsfw2}, and DOVER~\citep{wu2023exploring}. The retained clips are classified into the four physical-process categories and rebalanced to reduce category dominance while preserving observable state evolution. Finally, Qwen3-VL-235B-A22B~\citep{bai2025qwen3} produces structured annotations that describe object properties, state changes, and causal event sequences rather than only surface-level captions. To ensure annotation quality, we employ a team of trained professional data annotators to randomly audit the VLM-generated annotations against the source videos and annotation requirements. Annotations that are inconsistent with the observed video content or fail to satisfy the annotation requirements are manually reviewed and refined.

\paragraph{Duration distribution and training use.}
After the above process, \name-120K contains 120{,}804 clips with broad coverage across temporal scales. As shown in Fig.~\ref{fig:dataset_overview}(c), most clips fall within 3--10\,s, with the largest bin at 5--10\,s, which suits learning local physical responses such as impacts, splashes, deformation, and recovery; at the same time, the dataset retains a portion of longer clips to cover slower processes such as melting, burning, diffusion, and mixing. This balance in duration allows \name-120K to supervise both short-horizon physical responses and multi-stage process evolution. In our framework, the dataset serves as the real-world physical prior for supervised fine-tuning: the teacher model internalizes how physical states evolve, after which these priors are distilled into the few-step causal generator and further aligned during causal rollout generation.

\begin{figure*}[t]
  \centering
  \includegraphics[width=0.85\linewidth]{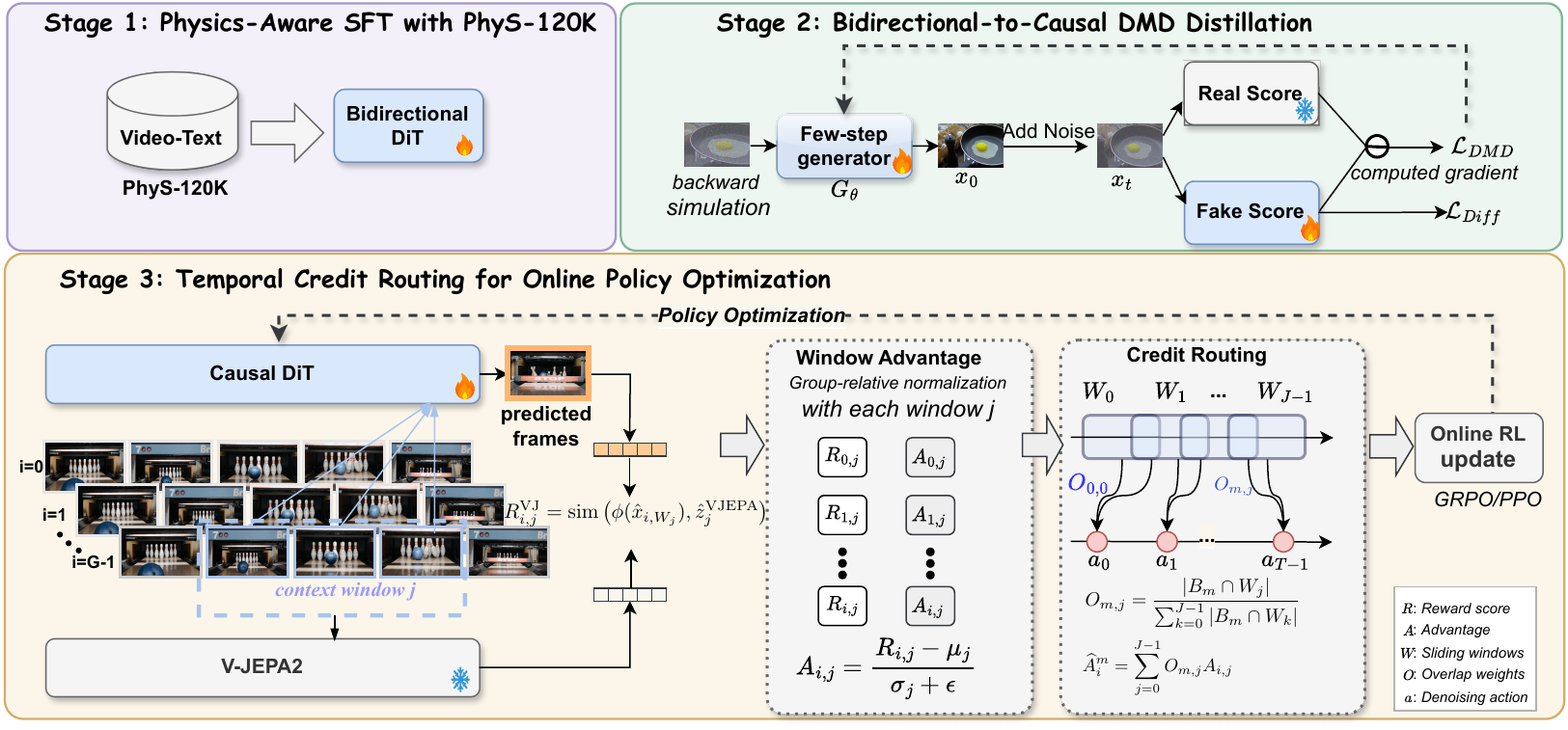}
  \caption{Overview of \name. Stages 1 and 2 transfer physical priors from a bidirectional teacher to a few-step causal generator. Stage 3 computes advantages $A_{i,j}$ over overlapping windows $W_j$ and routes them to the denoising actions.
  }
  \label{fig:method_overview}
\end{figure*}

\section{Method}
\label{sec:method}

\subsection{Overview and Problem Setup}
\label{sec:method-overview}

We develop \name as a three-stage framework for adapting a pretrained video generator to physically plausible streaming generation.
First, we fine-tune a bidirectional video DiT on \name-120K to acquire priors over real physical processes.
Second, we distill this teacher into a few-step causal generator that produces future frames block by block.
Third, we optimize the causal generator with \emph{Temporal Credit Routing} (TCR), which routes overlapping-window feedback to temporally aligned denoising actions.
Fig.~\ref{fig:method_overview} summarizes the framework.

Let $\mathbf{x}_{c}$ denote the observed context frames and $\mathbf{c}$ denote the text condition.
The model generates a future video $\mathbf{x}^{1:N}$ containing $N$ frames.
We partition these frames into $M$ consecutive causal blocks
$\{B_m\}_{m=1}^{M}$, where
$B_m \subseteq \{1,\ldots,N\}$ is the frame-index set of block $m$,
$\bigcup_{m=1}^{M} B_m=\{1,\ldots,N\}$,
and $B_{<m}=\bigcup_{q<m}B_q$.
The causal generator $G_\theta$ induces the blockwise distribution
\begin{equation}
p_\theta\!\left(
\mathbf{x}^{1:N}
\mid
\mathbf{x}_{c},\mathbf{c}
\right)
=
\prod_{m=1}^{M}
p_\theta\!\left(
\mathbf{x}^{B_m}
\mid
\mathbf{x}_{c},
\mathbf{x}^{B_{<m}},
\mathbf{c}
\right).
\label{eq:causal_factorization}
\end{equation}

Each block is produced through a small number of stochastic denoising transitions.
We use $\pi_\theta$ to denote the transition policy induced by $G_\theta$.
Standard video-level policy optimization assigns the same group-relative advantage to all transitions in all blocks.
Such a uniform signal cannot distinguish a locally implausible event from otherwise plausible portions of the rollout.
TCR addresses this temporal coarseness by aggregating overlapping-window advantages for each aligned denoising action.

\subsection{From Physical Priors to a Streaming Policy}
\label{sec:physical_streaming}

\paragraph{Physics-aware supervised fine-tuning.}
\label{sec:stage1}
We first adapt the bidirectional Wan2.1-14B to real physical processes.
Each clip in \name-120K is paired with a physics-aware caption describing observable object attributes, state changes, and temporal relations among events.
We retain the teacher's original flow-matching objective and fine-tune it with LoRA on these video--caption pairs.
This stage yields a physics-aware bidirectional teacher $\theta_{\mathrm{tea}}$, which provides the target distribution for the subsequent causal distillation stage.

\paragraph{Few-step causal distillation.}
\label{sec:stage2}
We distill the physics-aware teacher $\theta_{\mathrm{tea}}$ into
a few-step causal generator $G_\theta$ using
DMD~\citep{yin2024dmd}.
Following Self Forcing~\citep{huang2025selfforcing}, the student
performs an autoregressive self-rollout under
Eq.~\eqref{eq:causal_factorization}, such that each generated block
$\hat{\mathbf{x}}_0^{B_m}$ is conditioned on its own history
$\hat{\mathbf{x}}_0^{B_{<m}}$.
Thus, the complete rollout $\hat{\mathbf{x}}_0$ is sampled from the
same causal distribution used at inference.
We sample a diffusion timestep $\tau$ and perturb the rollout as
$\hat{\mathbf{x}}_\tau
=\alpha_\tau\hat{\mathbf{x}}_0+\sigma_\tau\boldsymbol{\eta}$,
where $\boldsymbol{\eta}\sim\mathcal{N}(\mathbf{0},\mathbf{I})$.

Let $p_{\theta,\tau}$ and $p_{\mathrm{tea},\tau}$ denote the noised
student-rollout and teacher distributions, respectively, with
conditioning omitted for brevity.
We optimize the holistic distribution-matching objective
\begin{equation}
\mathcal{L}_{\mathrm{DMD}}(\theta)
=
\mathbb{E}_{\tau}
\left[
D_{\mathrm{KL}}
\left(
p_{\theta,\tau}
\,\Vert\,
p_{\mathrm{tea},\tau}
\right)
\right].
\label{eq:dmd_objective}
\end{equation}
The frozen teacher score $s_{\theta_{\mathrm{tea}}}$ and the online
generator score $s_{\mathrm{gen},\xi}$ correspond to the Real Score
and Fake Score in Fig.~\ref{fig:method_overview}, respectively, while
$s_{\mathrm{gen},\xi}$ is updated using the standard flow-matching
objective.

\begin{table*}[!t]
\centering
\caption{Main results on four physics-aware video generation benchmarks. Bidirectional generations are 5 seconds at 40 NFE, and causal generations are 10 seconds at 4--5 NFE. Parenthetical percentages show relative changes computed from the displayed scores, using Wan2.1-14B as the baseline for SFT and each original Forcing method as the baseline for full \name.}
\label{tab:main_physics_results}
\fontsize{6.5}{8.5}\selectfont
\setlength{\tabcolsep}{0.45pt}
\renewcommand{\arraystretch}{1.0}
\begin{tabular}{lc*{11}{r@{\hspace{0.65pt}}l}}
\toprule
\multirow{2}{*}{\textbf{Method}} & \multirow{2}{*}{\textbf{NFE}} & \multicolumn{2}{c}{\textbf{PhysicsIQ}} & \multicolumn{4}{c}{\textbf{VideoPhy}} & \multicolumn{6}{c}{\textbf{VideoPhy2}} & \multicolumn{10}{c}{\textbf{PhyGenBench}} \\
\cmidrule(lr){3-4}\cmidrule(lr){5-8}\cmidrule(lr){9-14}\cmidrule(lr){15-24}
 &  & \multicolumn{2}{c}{\textbf{Score$\uparrow$}} & \multicolumn{2}{c}{\textbf{PC$\uparrow$}} & \multicolumn{2}{c}{\textbf{SA$\uparrow$}} & \multicolumn{2}{c}{\textbf{PC$\uparrow$}} & \multicolumn{2}{c}{\textbf{SA$\uparrow$}} & \multicolumn{2}{c}{\textbf{Joint$\uparrow$}} & \multicolumn{2}{c}{\textbf{Mechanics$\uparrow$}} & \multicolumn{2}{c}{\textbf{Optics$\uparrow$}} & \multicolumn{2}{c}{\textbf{Thermal$\uparrow$}} & \multicolumn{2}{c}{\textbf{Material$\uparrow$}} & \multicolumn{2}{c}{\textbf{Average$\uparrow$}} \\
\midrule
Wan2.1-1.3B & 40 & -- &  & 24.1 &  & 53.8 &  & 55.0 &  & 29.1 &  & 23.9 &  & 0.383 &  & 0.473 &  & 0.300 &  & 0.325 &  & 0.381 &  \\
Wan2.1-14B & 40 & 26.9 &  & 24.4 &  & 54.7 &  & 74.2 &  & 54.4 &  & 46.0 &  & 0.383 &  & 0.513 &  & 0.367 &  & 0.300 &  & 0.399 &  \\
\rowcolor{tcrrowblue}
\textbf{+ \name-120K SFT} & 40 & \textbf{31.8} & \tcrpct{+18.2} & \textbf{28.3} & \tcrpct{+16.0} & \textbf{55.5} & \tcrpct{+1.5} & \textbf{79.8} & \tcrpct{+7.5} & \textbf{56.2} & \tcrpct{+3.3} & \textbf{51.1} & \tcrpct{+11.1} & \textbf{0.439} & \tcrpct{+14.6} & \textbf{0.668} & \tcrpct{+30.2} & \textbf{0.382} & \tcrpct{+4.1} & \textbf{0.331} & \tcrpct{+10.3} & \textbf{0.473} & \tcrpct{+18.5} \\
\midrule
Self Forcing & 4 & 16.9 &  & 18.9 &  & 50.0 &  & 54.7 &  & 24.6 &  & 24.4 &  & 0.347 &  & \textbf{0.455} &  & 0.316 &  & 0.319 &  & 0.368 &  \\
\quad + SFT Teacher & 4 & 17.4 &  & 20.6 &  & 52.7 &  & 56.8 &  & 24.9 &  & 23.6 &  & 0.377 &  & 0.428 &  & 0.328 &  & 0.302 &  & 0.365 &  \\
\quad + Flow-GRPO & 4 & 18.2 &  & 20.1 &  & 45.5 &  & 52.7 &  & 21.0 &  & 19.3 &  & 0.383 &  & 0.347 &  & 0.311 &  & 0.310 &  & 0.340 &  \\
\rowcolor{tcrrowblue}
\quad + TCR (Ours) & 4 & \textbf{20.9} & \tcrpct{+23.7} & \textbf{22.4} & \tcrpct{+18.5} & \textbf{53.7} & \tcrpct{+7.4} & \textbf{62.7} & \tcrpct{+14.6} & \textbf{33.8} & \tcrpct{+37.4} & \textbf{26.9} & \tcrpct{+10.2} & \textbf{0.394} & \tcrpct{+13.5} & 0.451 & \tcrpct{-0.9} & \textbf{0.368} & \tcrpct{+16.5} & \textbf{0.322} & \tcrpct{+0.9} & \textbf{0.389} & \tcrpct{+5.7} \\
\hdashline
Rolling Forcing & 5 & 16.2 &  & 26.2 &  & 48.0 &  & 67.5 &  & 26.2 &  & 23.0 &  & 0.350 &  & 0.470 &  & 0.333 &  & 0.305 &  & 0.373 &  \\
\quad + SFT Teacher & 5 & 17.9 &  & 27.0 &  & 49.6 &  & 69.7 &  & 27.5 &  & 23.9 &  & 0.371 &  & 0.467 &  & 0.346 &  & 0.314 &  & 0.382 &  \\
\quad + Flow-GRPO & 5 & 17.1 &  & 26.3 &  & 47.3 &  & 71.7 &  & 24.9 &  & 22.8 &  & 0.377 &  & 0.476 &  & 0.352 &  & \textbf{0.320} &  & 0.389 &  \\
\rowcolor{tcrrowblue}
\quad + TCR (Ours) & 5 & \textbf{18.6} & \tcrpct{+14.8} & \textbf{28.0} & \tcrpct{+6.9} & \textbf{52.2} & \tcrpct{+8.8} & \textbf{76.4} & \tcrpct{+13.2} & \textbf{28.9} & \tcrpct{+10.3} & \textbf{24.7} & \tcrpct{+7.4} & \textbf{0.382} & \tcrpct{+9.1} & \textbf{0.608} & \tcrpct{+29.4} & \textbf{0.374} & \tcrpct{+12.3} & 0.197 & \tcrpct{-35.4} & \textbf{0.405} & \tcrpct{+8.6} \\
\hdashline
Causal Forcing & 4 & 10.2 &  & 21.5 &  & 42.4 &  & 54.8 &  & 24.2 &  & 19.8 &  & 0.349 &  & 0.433 &  & 0.322 &  & 0.317 &  & 0.362 &  \\
\quad + SFT Teacher & 4 & 11.8 &  & 22.6 &  & 44.0 &  & 58.1 &  & 25.3 &  & 20.7 &  & 0.357 &  & 0.437 &  & 0.331 &  & 0.320 &  & 0.368 &  \\
\quad + Flow-GRPO & 4 & 11.4 &  & 20.4 &  & 39.1 &  & 60.5 &  & 22.1 &  & 18.5 &  & 0.363 &  & 0.287 &  & 0.339 &  & 0.324 &  & 0.325 &  \\
\rowcolor{tcrrowblue}
\quad + TCR (Ours) & 4 & \textbf{13.4} & \tcrpct{+31.4} & \textbf{26.1} & \tcrpct{+21.4} & \textbf{47.8} & \tcrpct{+12.7} & \textbf{68.1} & \tcrpct{+24.3} & \textbf{25.7} & \tcrpct{+6.2} & \textbf{24.4} & \tcrpct{+23.2} & \textbf{0.376} & \tcrpct{+7.7} & \textbf{0.491} & \tcrpct{+13.4} & \textbf{0.361} & \tcrpct{+12.1} & \textbf{0.328} & \tcrpct{+3.5} & \textbf{0.397} & \tcrpct{+9.7} \\
\bottomrule
\end{tabular}
\end{table*}

\subsection{Temporal Credit Routing}
\label{sec:tcr}
\label{sec:stage3}

Physical violations often occupy only a short interval of a generated video.
A global rollout reward can indicate whether the complete video is plausible, but it does not reveal where a violation occurs.
TCR constructs a temporally local learning signal through three operations:
window-level physical scoring, per-window group normalization, and overlap-based routing.

\paragraph{Window-level physical feedback.}
For each condition $(\mathbf{x}_{c},\mathbf{c})$, the behavior policy
$\pi_{\theta_{\mathrm{old}}}$ samples a group of $G$ future rollouts
$\{\hat{\mathbf{x}}^{i}\}_{i=0}^{G-1}$.
We cover the generated frames with $J$ overlapping temporal windows
$\{W_j\}_{j=0}^{J-1}$, where
$W_j\subseteq\{1,\ldots,N\}$ and
$\bigcup_{j=0}^{J-1}W_j=\{1,\ldots,N\}$.

TCR is agnostic to the particular window-level reward model.
Let $R_\phi$ denote a frozen scorer that evaluates the physical consistency of a generated window given the context window $\mathcal{C}_j$ shown in Fig.~\ref{fig:method_overview}:
\begin{equation}
R_{i,j}
=
R_\phi\!\left(
\hat{\mathbf{x}}^{i}_{W_j}
\mid
\mathcal{C}_j,\mathbf{c}
\right),
\qquad
A_{i,j}
=
\frac{R_{i,j}-\mu_j}{\sigma_j+\epsilon},
\label{eq:window_advantage}
\end{equation}
where
\begin{equation}
\mu_j
=
\frac{1}{G}\sum_{i=0}^{G-1}R_{i,j},
\qquad
\sigma_j
=
\sqrt{
\frac{1}{G}
\sum_{i=0}^{G-1}
\left(R_{i,j}-\mu_j\right)^2
}.
\label{eq:window_statistics}
\end{equation}
Here, $\epsilon>0$ is a numerical stabilizer.
Normalizing rewards across the $G$ rollouts separately for each window prevents intrinsically easy or difficult temporal regions from dominating the policy update.

In our implementation, $R_\phi$ uses a frozen V-JEPA2 predictor~\citep{assran2025v} to compare the future representation predicted from $\mathcal{C}_j$ with that encoded from $W_j$.
To encourage sufficient motion in generated videos, we introduce an optical-flow reward computed by RAFT~\citep{teed2020raft}, whose contribution is controlled by $\lambda_{\mathrm{flow}}$.
The composite reward is
$R_{i,j}=R^{\mathrm{VJ}}_{i,j}+\lambda_{\mathrm{flow}}R^{\mathrm{Flow}}_{i,j}$.

\paragraph{Weighted Credit Routing.}
Reward windows $\{W_j\}_{j=0}^{J-1}$ slide over the generated frame sequence.
For each causal block $B_m$, we aggregate the advantages of all reward windows that overlap its generated frames:
\begin{equation}
O_{m,j}
=
\frac{
\left|B_m\cap W_j\right|
}{
\sum_{k=0}^{J-1}\left|B_m\cap W_k\right|
},
\qquad
\widehat{A}_{i}^{m}
=
\sum_{j=0}^{J-1}O_{m,j}A_{i,j}.
\label{eq:tcr_routing}
\end{equation}
The normalized weights satisfy $\sum_jO_{m,j}=1$.
The resulting block-level advantage $\widehat{A}_{i}^{m}$ is shared by all denoising transitions that generate block $B_m$.

\paragraph{Policy optimization.}
We optimize the causal generator using the stochastic transition policy introduced by Flow-GRPO~\citep{liu2025flowgrpo}.
Let $m\in\{1,\ldots,M\}$ index causal blocks and $s\in\{1,\ldots,S\}$ index the denoising transitions within each block.
For rollout $i$, block $m$, and transition $s$, let
$\mathbf{u}_{i,m,s}$ collect the current noisy state, generated history, observed context, and text condition.
Let $a_{i,m,s}$ denote the sampled transition.
The importance ratio is
\begin{equation}
r_{i,m,s}(\theta)
=
\frac{
\pi_\theta
\left(
a_{i,m,s}\mid\mathbf{u}_{i,m,s}
\right)
}{
\pi_{\theta_{\mathrm{old}}}
\left(
a_{i,m,s}\mid\mathbf{u}_{i,m,s}
\right)
}.
\label{eq:importance_ratio}
\end{equation}

We replace the global rollout advantage with the overlap-routed advantage $\widehat{A}_{i}^{m}$.
The resulting per-transition surrogate is
\begin{equation}
\begin{aligned}
\bar r_{i,m,s}(\theta)
&=
\operatorname{clip}\!\left(
r_{i,m,s}(\theta),
1-\varepsilon_{\mathrm{clip}},
1+\varepsilon_{\mathrm{clip}}
\right),\\
\ell_{i,m,s}^{\mathrm{TCR}}(\theta)
&=
\min\!\left\{
r_{i,m,s}(\theta)\widehat{A}_{i}^{m},
\bar r_{i,m,s}(\theta)\widehat{A}_{i}^{m}
\right\}.
\end{aligned}
\label{eq:tcr_surrogate}
\end{equation}
where $\varepsilon_{\mathrm{clip}}$ is the clipping threshold.
We define the corresponding per-transition KL penalty as
\begin{equation}
\mathcal{K}_{i,m,s}(\theta)
=
D_{\mathrm{KL}}\!\left(
\pi_\theta(\cdot\mid\mathbf{u}_{i,m,s})
\,\Vert\,
\pi_{\mathrm{ref}}(\cdot\mid\mathbf{u}_{i,m,s})
\right),
\label{eq:transition_kl}
\end{equation}
where $\pi_{\mathrm{ref}}$ is the frozen reference policy.
The final objective is
\begin{equation}
\resizebox{0.98\columnwidth}{!}{$\displaystyle
\theta^\star
=
\arg\max_{\theta}\,
\mathbb{E}\!\left[
\frac{1}{GM}
\sum_{i=0}^{G-1}
\sum_{m=1}^{M}
\frac{1}{S}
\sum_{s=1}^{S}
\left(
\ell_{i,m,s}^{\mathrm{TCR}}(\theta)
-
\beta\mathcal{K}_{i,m,s}(\theta)
\right)
\right]
$}
\label{eq:tcr_obj}
\end{equation}
where $\beta$ controls KL regularization.

\begin{table}[t]
\centering
\caption{Long-horizon rollout results at 15 and 20 seconds.}
\label{tab:long_horizon}
\footnotesize
\setlength{\tabcolsep}{1.55pt}
\renewcommand{\arraystretch}{1.0}
\begin{tabular}{@{}lccccc@{}}
\toprule
\multirow{2}{*}{\textbf{Method}}
& \multicolumn{3}{c}{\textbf{VBench-Long $\uparrow$}}
& \multicolumn{2}{c}{\textbf{VideoPhy $\uparrow$}} \\
\cmidrule(lr){2-4}\cmidrule(lr){5-6}
& \metricheading{Temporal}{Quality}
& \metricheading{Dynamic}{Degree}
& \metricheading{Visual}{Stability}
& \metricheading{Physical}{Commonsense}
& \metricheading{Semantic}{Adherence} \\
\midrule
\multicolumn{6}{c}{\textbf{\textit{Results on 15s}}} \\
\midrule
Self Forcing & 89.75 & 46.37 & 53.61 & 18.4 & 47.3 \\
\textbf{+ \name} & \textbf{94.41} & \textbf{53.25} & \textbf{58.93} & \textbf{20.4} & \textbf{49.2} \\
\hdashline
Rolling Forcing & 92.91 & 57.39 & 56.83 & 23.7 & 43.4 \\
\textbf{+ \name} & \textbf{94.07} & \textbf{65.53} & \textbf{60.71} & \textbf{26.3} & \textbf{49.8} \\
\midrule
\multicolumn{6}{c}{\textbf{\textit{Results on 20s}}} \\
\midrule
Self Forcing & 89.30 & 37.92 & 43.56 & 16.6 & 43.8 \\
\textbf{+ \name} & \textbf{91.82} & \textbf{48.29} & \textbf{47.01} & \textbf{19.7} & \textbf{47.8} \\
\hdashline
Rolling Forcing & 90.47 & 46.39 & 52.63 & 21.7 & 38.6 \\
\textbf{+ \name} & \textbf{92.36} & \textbf{58.23} & \textbf{52.71} & \textbf{25.9} & \textbf{47.4} \\
\bottomrule
\end{tabular}
\end{table}

\section{Experiments}
\label{sec:exp}

\subsection{Experimental Setup}
\label{sec:exp-setup}

\begin{figure*}[!b]
  \centering
  \includegraphics[width=0.92\textwidth]{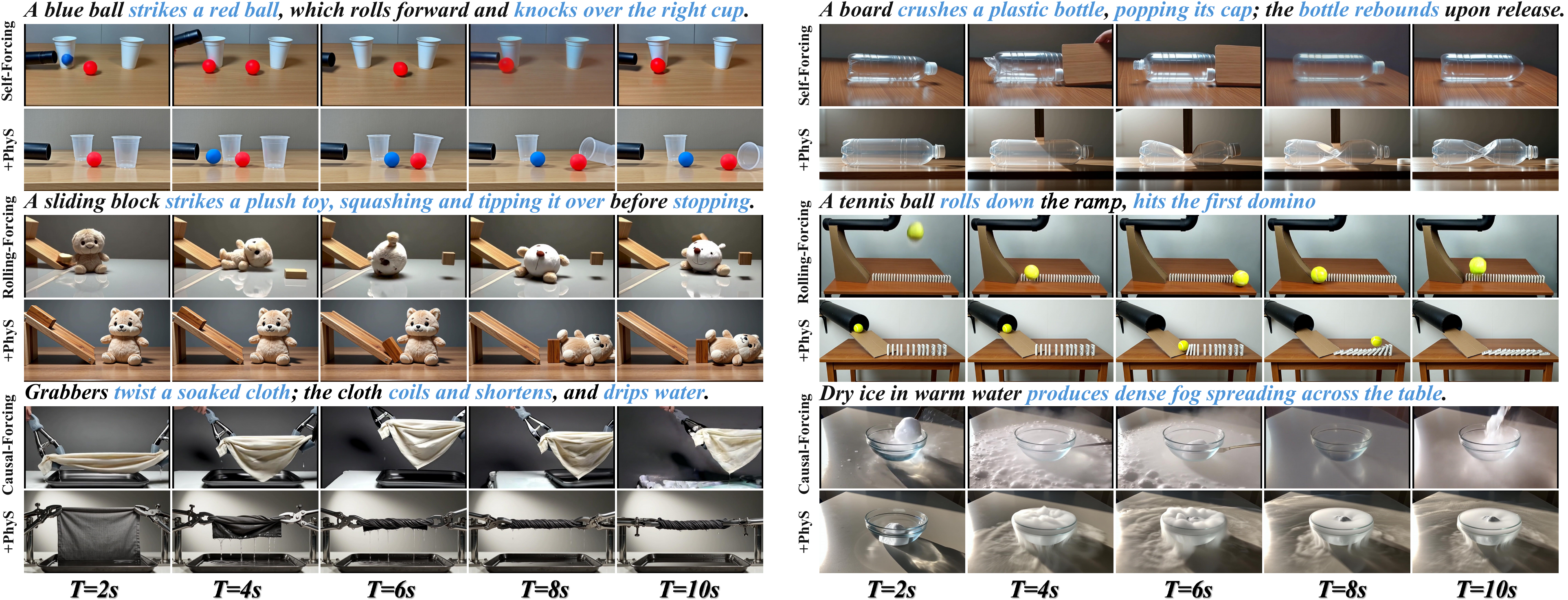}
  \caption{Long-horizon causal rollouts. Each pair shows the Forcing baseline (top) and its \name counterpart (bottom) for the same prompt. \name better preserves visual realism and physical plausibility throughout the rollout.}
  \label{fig:long_horizon_comparison}
\end{figure*}

\paragraph{Implementation Details.}
We fine-tune the T2V and I2V versions of Wan2.1-14B~\citep{wan2025wan} on \name-120K, then use DMD~\citep{yin2024dmd} to distill the teachers into few-step Wan2.1-1.3B causal students under the official Self, Rolling, and Causal Forcing configurations. During TCR, we freeze the base model and train LoRA adapters ($r=32$, $\alpha=64$) with rollout group size $G=8$ and KL coefficient $\beta=0.05$. The reward adds a RAFT optical-flow term weighted by $2.0$ to the V-JEPA2 consistency score.

\paragraph{Evaluation Protocols.}
We evaluate 5-second videos from bidirectional teachers at 40 NFE and 10-second causal rollouts at 4--5 NFE. We evaluate 10 videos per benchmark prompt, each generated with a distinct random seed. PhysicsIQ~\citep{motamed2026generative} tests context-conditioned I2V. For T2V, VideoPhy~\citep{Bansal2025} and VideoPhy2~\citep{bansal2025videophy} report physical commonsense (PC) and semantic adherence (SA). VideoPhy2 also reports their joint success rate. PhyGenBench~\citep{meng2024towards} assesses physical plausibility across mechanics, optics, thermal phenomena, and material interactions.

\subsection{Main Results}
\label{sec:exp-main}

\textbf{Injecting physical priors into the teacher.} At the same 40 NFE, SFT on \name-120K improves the bidirectional Wan2.1-14B teacher on every metric in Table~\ref{tab:main_physics_results}. PhysicsIQ, VideoPhy PC, and VideoPhy2 Joint rise by 18.2\%, 16.0\%, and 11.1\%, respectively, while the average PhyGenBench score improves by 18.5\% over the original teacher.

\begin{figure*}[!t]
  \centering
  \includegraphics[width=0.92\textwidth]{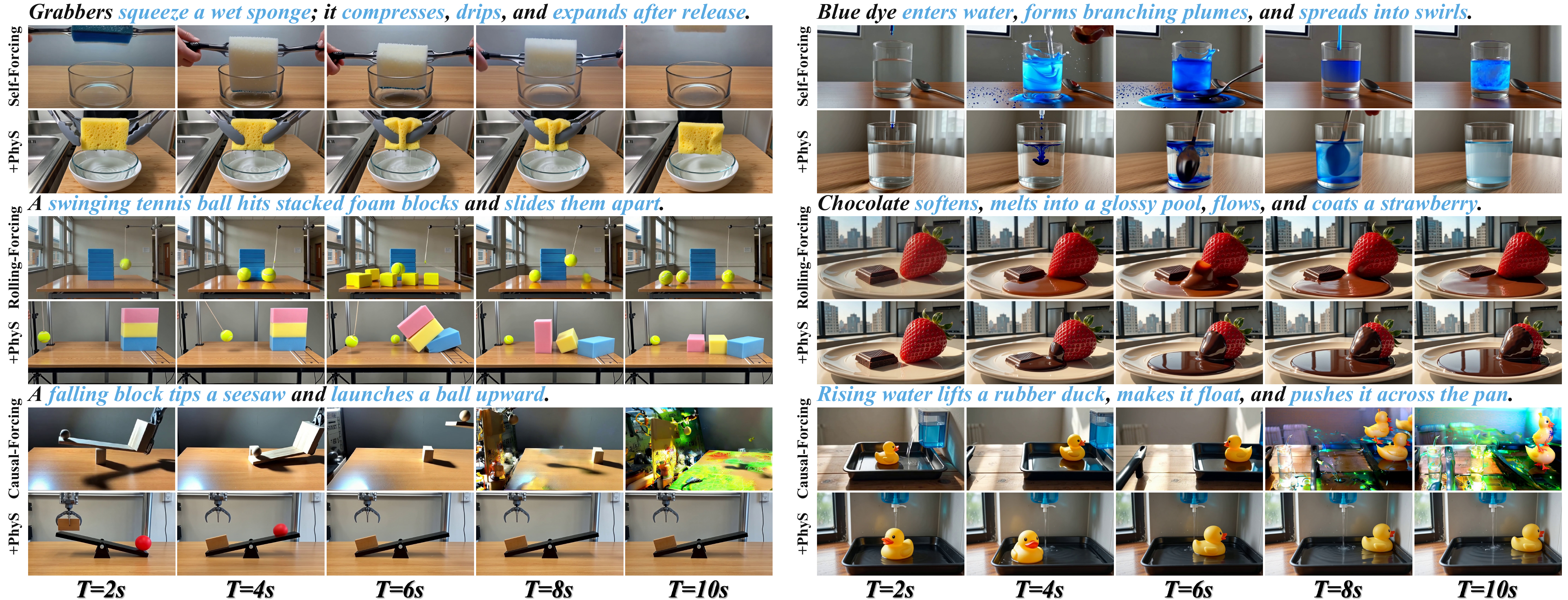}
  \caption{Additional 10-second rollouts. Two examples per Forcing method compare the baseline (top) with \name (bottom).}
  \label{fig:long_horizon_case_study}
\end{figure*}

\textbf{Transferring physical priors to causal students.} Under the 10 seconds streaming-rollout protocol for few-step causal DiTs, distillation from the SFT teacher improves PhysicsIQ from 16.9 to 17.4 for Self Forcing, 16.2 to 17.9 for Rolling Forcing, and 10.2 to 11.8 for Causal Forcing. Most VideoPhy and VideoPhy2 PC and SA metrics also improve, confirming that the acquired physical priors transfer across causal distillation strategies. However, several regressions indicate that causal distillation does not preserve these priors uniformly.

\textbf{Online alignment with physical rewards.} At the same budget of 4--5 NFEs, the full \name pipeline improves PhysicsIQ by 23.7\%, 14.8\%, and 31.4\% over Self Forcing, Rolling Forcing, and Causal Forcing, respectively, and also improves VideoPhy PC and VideoPhy2 Joint. With the joint RAFT--V-JEPA2 reward, TCR outperforms the corresponding video-level Flow-GRPO variants by 8.8\%--17.5\% on PhysicsIQ and by 8.3\%--39.4\% on VideoPhy2 Joint. These results support routing local reward signals to the causal blocks that overlap the scored windows. On PhyGenBench, the full \name pipeline raises the average score by 5.7\%, 8.6\%, and 9.7\% over the three original baselines, respectively, and surpasses each Flow-GRPO counterpart. All three backbones improve in Mechanics and Thermal, and Causal Forcing improves in all four categories. The only declines relative to the original baselines are in Optics for Self Forcing and in Material for Rolling Forcing.

\subsection{Long-Horizon Streaming Rollout}
\label{sec:exp-streaming}

Table~\ref{tab:long_horizon} reports three long-horizon metrics from VBench-Long~\citep{huang2024vbench,huang2026vbenchpp} (Temporal Quality, Dynamic Degree, and Visual Stability), along with VideoPhy PC and SA. \name improves all five metrics over Self Forcing and Rolling Forcing for both 15- and 20-second rollouts. For 15-second rollouts, \name improves Dynamic Degree by 6.88 and 8.14 points over Self Forcing and Rolling Forcing, respectively, and VideoPhy PC by 2.0 and 2.6 points.

Figs.~\ref{fig:long_horizon_comparison} and~\ref{fig:long_horizon_case_study} illustrate how physical effects evolve across causal blocks. In Fig.~\ref{fig:long_horizon_comparison}, the baselines often render the initiating event but fail to capture its later consequences, such as a collision setting nearby objects in motion or fog continuing to spread. Fig.~\ref{fig:long_horizon_case_study} shows similar failures in scenes involving deformation, melting, and buoyancy. Across these examples, \name better captures both the initiating events and their downstream consequences, producing streaming rollouts that are more physically plausible and causally consistent.

\subsection{Human Study}
\label{sec:human-study}

\begin{figure}[!t]
  \centering
  \includegraphics[width=\linewidth,trim=15.03bp 0bp 16.55bp 0bp,clip]{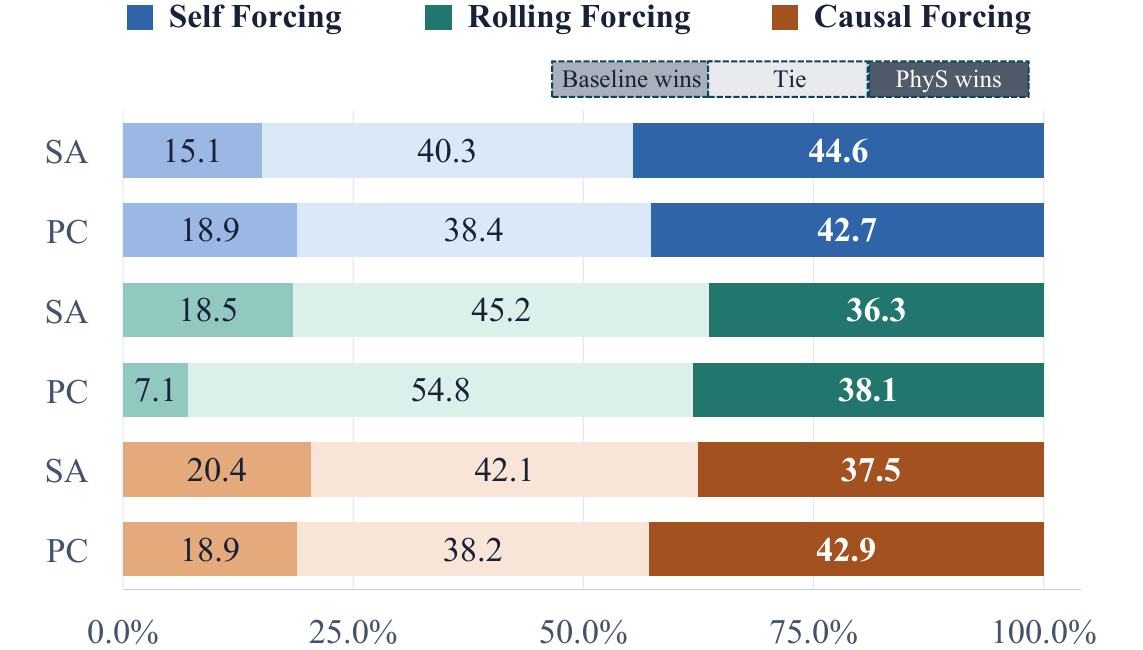}
  \caption{Human rater judgments for \name vs.\ Forcing baselines on semantic adherence (SA) and physical commonsense (PC).}
  \label{fig:human_study}
\end{figure}

We use the GPT API to generate prompts from predefined scenarios and keywords, and retain 600 prompts after manual screening. The set contains 100 prompts for each of the four \name-120K categories, plus 100 each for Physical Activities and Robotics Action, two categories outside \name-120K that test generalization. For each prompt, we generate a 10-second video with each baseline and with its \name counterpart.
\begin{table}[!t]
\centering
\caption{Data and annotation ablation.}
\label{tab:data_annotation_ablation}
\small
\setlength{\tabcolsep}{1.4pt}
\renewcommand{\arraystretch}{1.0}
\begin{tabular}{@{}lccccc@{}}
\toprule
\multirow{2}{*}{\textbf{Setting}}
& \multirow{2}{*}{\shortstack{\textbf{Annotation}\\\textbf{Type}}}
& \multicolumn{3}{c}{\textbf{VBench $\uparrow$}}
& \textbf{PhysicsIQ $\uparrow$} \\
\cmidrule(lr){3-5}\cmidrule(lr){6-6}
&
& \textbf{Quality}
& \textbf{Semantic}
& \textbf{Total}
& \textbf{Score} \\
\midrule
Wan2.1-14B
& -- & 83.1 & 79.5 & 82.4 & 26.9 \\
General Data
& Desc. & 83.4 & 77.1 & 82.1 & 25.2 \\
WISA-80K
& Desc.+Prop. & 81.5 & 80.6 & 81.3 & 27.9 \\
\name-80K
& Desc. & 84.2 & 80.9 & 83.5 & 27.3 \\
\textbf{\name-80K}
& \textbf{Full} & \textbf{87.1} & \textbf{82.7} & \textbf{86.2} & \textbf{31.4} \\
\bottomrule
\end{tabular}
\end{table}
Twelve raters compare each video pair on SA and PC and, for each criterion, select the preferred video or a tie. For each baseline and criterion, we report the proportions of wins, ties, and losses over all individual judgments. As shown in Fig.~\ref{fig:human_study}, \name receives more favorable judgments than each Forcing baseline on both criteria. The largest gap is on PC against Rolling Forcing: 38.1\% of judgments favor \name, 7.1\% favor the baseline, and 54.8\% are ties.

\subsection{Ablation Studies}
\label{sec:exp-ablation}
\label{sec:exp-tcr}

\textbf{Dataset and annotation effectiveness.} Table~\ref{tab:data_annotation_ablation} compares how data sources and annotation types affect physical-prior injection under a matched scale. All SFT variants use the same Wan2.1-14B backbone and training recipe with approximately 80K clips, so the observed differences do not arise simply from additional training data. General Data uses generic descriptions (Desc.) for 80K videos sampled from open-source video datasets. WISA-80K~\citep{wang2025wisa} covers 17 physical phenomena and augments descriptions with physical properties (Desc.+Prop.). Our \name-80K variants use the same process-centric videos but differ in annotation type: the generic version uses descriptions only (Desc.), while the full structured version (Full) additionally includes physical properties, temporally ordered state changes, and causal relations. General Data provides no consistent improvement over the pretrained baseline, while WISA-80K improves Semantic and PhysicsIQ, demonstrating the benefit of physics-oriented data and explicit physical attributes. Replacing generic descriptions with Full annotations on the same \name-80K videos improves all four metrics, demonstrating the benefit of structured physical supervision.

\begin{figure}[t]
  \centering
  \includegraphics[width=\linewidth]{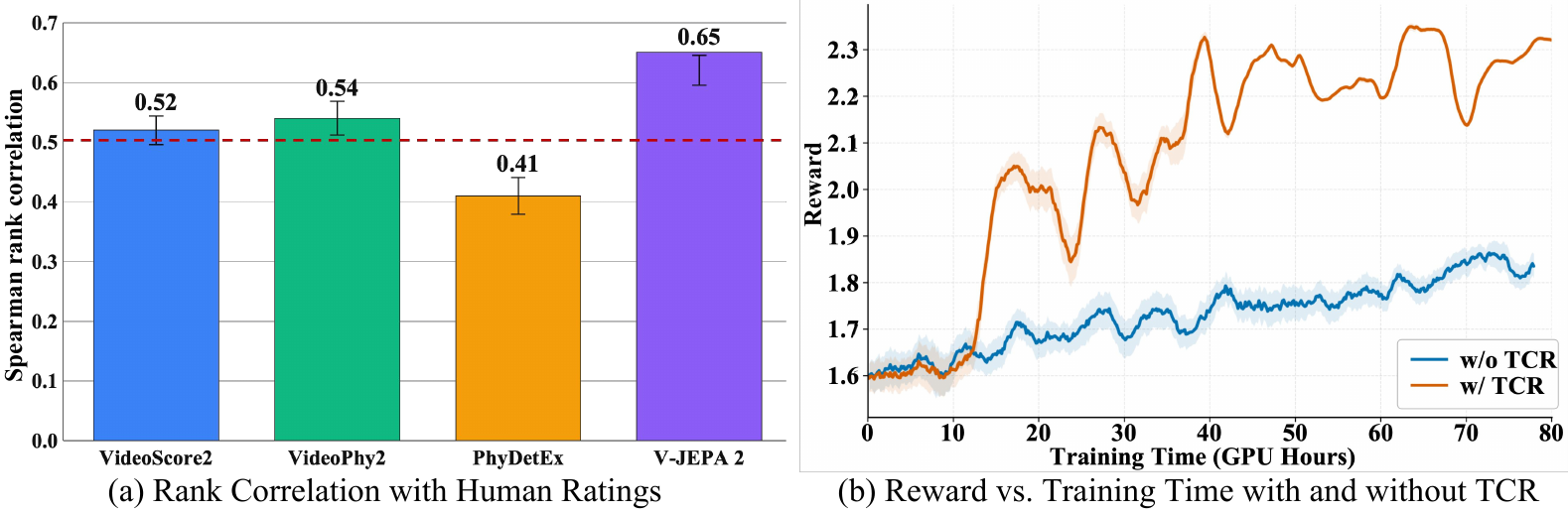}
  \caption{(a) Spearman rank correlation with human ratings. \mbox{V-JEPA2} ranks highest ($\rho = 0.65$). (b) TCR achieves higher training rewards at comparable GPU-hour budgets.}
  \label{fig:reward_comparison}
\end{figure}

\textbf{Temporal credit routing and reward models.}
Fig.~\ref{fig:reward_comparison}(a) compares four reward models against human ratings of physical plausibility: VideoScore2~\citep{he2025videoscore2}, VideoPhy2~\citep{bansal2025videophy}, \mbox{PhyDetEx}~\citep{wang2025phydetex}, and V-JEPA2~\citep{assran2025v}. We randomly sample 1,000 prompts from our training set in proportion to its category distribution. For each prompt, the same generator, either Wan2.1-14B or Wan2.2-5B, produces both videos. For each reward model, we independently rank its scores and the human scores for the same videos, assign average ranks to ties, and compute the Pearson correlation between the two sets of ranks. V-JEPA2 achieves the highest Spearman rank correlation ($\rho = 0.65$) and serves as the physical-consistency component of our reward.

\begin{table}[t]
\centering
\caption{Ablations of temporal credit routing and reward models for 10-second Self Forcing rollouts. TQ, DD, and VS denote Temporal Quality, Dynamic Degree, and Visual Stability.}
\label{tab:reward_tcr_ablation}
\fontsize{7.5}{9.5}\selectfont
\setlength{\tabcolsep}{3pt}
\renewcommand{\arraystretch}{1.0}
\begin{tabular*}{\columnwidth}{@{\extracolsep{\fill}}lcccccc@{}}
\toprule
\multirow{2}{*}{\textbf{Variant}}
& {\fontsize{6.5}{8}\selectfont\textbf{PhysicsIQ $\uparrow$}}
& \multicolumn{3}{c}{\textbf{VBench-Long $\uparrow$}}
& \multicolumn{2}{c}{\textbf{VideoPhy $\uparrow$}} \\
\cmidrule(lr){2-2}\cmidrule(lr){3-5}\cmidrule(lr){6-7}
& {\fontsize{6.5}{8}\selectfont\textbf{Score}} & \textbf{TQ} & \textbf{DD} & \textbf{VS}
& \textbf{PC}
& \textbf{SA} \\
\midrule
Baseline
& 16.9 & 89.66 & 39.10 & 53.92 & 18.9 & 50.0 \\
SFT Teacher
& 17.4 & 90.26 & 46.49 & 51.62 & 20.6 & 52.7 \\
TCR
& 16.7 & 84.32 & 42.21 & 50.24 & 19.2 & 51.5 \\
\midrule
Flow-GRPO
& 18.2 & 82.37 & 43.20 & 54.24 & 20.1 & 45.5 \\
Window reward
& 19.6 & 89.34 & 44.35 & 56.26 & 21.3 & 48.6 \\
\midrule
RAFT
& 14.2 & 91.89 & 50.28 & 44.31 & 18.2 & 47.6 \\
V-JEPA2
& 19.4 & 92.52 & 49.14 & 57.37 & 22.1 & 52.6 \\
\textbf{TCR Full}
& \textbf{20.9} & \textbf{96.83} & \textbf{56.73} & \textbf{62.86} & \textbf{22.4} & \textbf{53.7} \\
\bottomrule
\end{tabular*}
\end{table}

Fig.~\ref{fig:reward_comparison}(b) compares TCR with vanilla GRPO, which broadcasts a single video-level reward across the rollout. Both runs start from the same Self Forcing student distilled from the SFT teacher and use identical hyperparameters. TCR achieves higher training rewards at \mbox{comparable GPU-hour budgets.}

Table~\ref{tab:reward_tcr_ablation} compares three groups of Self Forcing variants. In the first group, Baseline and SFT Teacher are students distilled from the original pretrained teacher and from the physics-aware SFT teacher, respectively, and TCR applies online alignment to Baseline. The second and third groups both start from the SFT Teacher student. The second group (two rows) varies reward routing while keeping the full reward and hyperparameters fixed. The third group (three rows) varies the reward model while keeping overlap routing and the training budget fixed. Applied to Baseline, TCR yields mixed results, whereas TCR Full achieves the best score on all six metrics.

\vspace{-8pt}
\section{Conclusion}

{\hyphenpenalty=10000 \exhyphenpenalty=10000 We introduce \name, a data and training framework that injects physical priors into existing Forcing pipelines for streaming video generation. Its dataset, \name-120K, provides real-world videos with structured annotations of object properties and causal state transitions. Using these annotations, we fine-tune a bidirectional 14B DiT teacher, distill its physical priors into a few-step 1.3B causal student, and further refine the student with online RL. In this RL stage, Temporal Credit Routing (TCR) decouples credit assignment from the Flow-GRPO objective by routing the reward of each scored window to the denoising actions of the causal blocks that overlap it. Across multiple Forcing baselines, \name improves the physical plausibility of long-horizon streaming rollouts.\par}

{\small
\bibliographystyle{ieeenat_fullname}
\bibliography{main}
}
\end{document}